\pdfoutput=1

\documentclass[11pt]{article}

\usepackage[final]{acl}

\usepackage{times}
\usepackage{latexsym}

\usepackage[T1]{fontenc}
\usepackage[utf8]{inputenc}

\usepackage{microtype}

\usepackage{inconsolata}

\usepackage{graphicx}
\usepackage{amsmath}
\usepackage{float}

\usepackage{color,soul}

\title{A Psycholinguistic Evaluation of Language Models' \\Sensitivity to Argument Roles}

\author{
 \textbf{Eun-Kyoung Rosa Lee\textsuperscript{1}},
 \textbf{Sathvik Nair\textsuperscript{1,2}},
 \textbf{Naomi H. Feldman\textsuperscript{1,2}}
\\
\\
 University of Maryland, College Park\\
 \textsuperscript{1} Department of Linguistics\\
 \textsuperscript{2} Institute for Advanced Computer Studies
\\
 \small{
   \textbf{Correspondence:} \href{mailto:ekleesla@umd.edu}{ekleesla@umd.edu}
 }
}

\begin{document}
\maketitle
\begin{abstract}
We present a systematic evaluation of large language models' sensitivity to argument roles, i.e., \textit{who} did what to \textit{whom}, by replicating psycholinguistic studies on human argument role processing. In three experiments, we find that language models are able to distinguish verbs that appear in plausible and implausible contexts, where plausibility is determined through the relation between the verb and its preceding arguments. However, none of the models capture the same selective patterns that human comprehenders exhibit during real-time verb prediction. This indicates that language models' capacity to detect verb plausibility does not arise from the same mechanism that underlies human real-time sentence processing. 
% Language models exhibit difficulty in computing complex relations between role-specified arguments and verbs which are not transparently represented in simple word co-occurrences.

\end{abstract}

\section{Introduction}

Humans rapidly make predictions when comprehending language. However, certain types of contextual information do not immediately impact predictions, and a well-studied case of this in the sentence processing literature involves argument roles.

Argument roles refer to the roles of participants that take part in the event described by a sentence, i.e., who is the agent (do-er of the action) and who is the patient (undergo-er of the action).
Extracting this information from the sentence and using it with prior knowledge to predict which event is being described is a hallmark of real-time language understanding.
For example, in (1a), the verb \textit{served} is a highly expected continuation given the preceding context, whereas swapping the argument roles, as in (1b), makes the same verb no longer appropriate.

\begin{enumerate}
    \item 
    \begin{enumerate}
        \item[a.] The customer that the waitress \textbf{served}
        \item[b.] The waitress that the customer \textbf{served}
    \end{enumerate}
\end{enumerate}
Surprisingly, studies with human participants have shown that the roles assigned to the arguments by the structure do not immediately impact verb prediction, in contrast to the context-independent lexical meanings of arguments. Human comprehenders show similar initial responses to a verb when it appears in role-appropriate and role-reversed contexts (e.g., 1a vs. 1b) \citep{kim2005independence,chow2016bag}. This has been taken to indicate that argument roles have a delayed impact on verb prediction in human sentence processing.

Recent work has used paradigms from experimental psycholinguistics to evaluate language models' representation of syntactic and semantic knowledge, and language models trained on next-word prediction alone have shown strong levels of correspondence with human behavioral and neural data. 
However, the extent to which they accurately encode and utilize structural information, such as argument roles, in relation to structure-independent word meanings, to determine sentence plausibility remains an open question.
Previous work has explored whether models can distinguish between plausible and implausible sentences involving argument role manipulations \citep{ettinger-2020-bert,papadimitriou-etal-2022-classifying-grammatical,wilson2023abstract,kauf2023event}. However, much of this research has focused on comparing full sentences rather than isolating the relationship between argument roles and the verb, often introducing confounding factors such as animacy. This makes it challenging to accurately assess models' sensitivity to argument role information.

In this paper, we take a new approach in evaluating role-sensitivity in large language models, by focusing on models' representations of verbs that appear in either plausible or implausible sentence contexts, where plausibility is determined based on the verb's compatibility with the preceding argument-role bindings. 
This approach draws insights from experimental work testing humans' role-sensitivity and therefore offers a more direct evaluation of language models' sensitivity to structural information in comparison to humans than previous studies.
Additionally, testing language models that are trained on next-word prediction provides a fertile testing ground for determining whether the systematic predictive patterns observed in human empirical behavior naturally arise from statistical co-occurrences and a prediction objective, as opposed to additional human cognitive processes. In this way, directly comparing predictive processing between humans and models can help us better understand the mechanisms that underlie human language processing.

% Language models that are trained on next-word prediction may be capable of learning relations between various arguments and verbs, i.e., between real-world events and participants that are likely to be involved in those events, based on co-occurrences between noun and verb combinations. However, to exhibit role-sensitivity at the verb, further generalizations must be made with regard to which role a participant likely takes on in a particular event, i.e., whether a waitress is often an agent or patient of a serving event. This knowledge must then be compared against the particular argument-role binding that is indicated by the structure of a given sentence. Therefore, examining whether language models can distinguish contextually plausible and implausible verbs, given the preceding argument-role bindings, serves as a rigorous test in examining models' sensitivity to argument role information and offers a more direct comparison with the human empirical patterns. 

We adapt materials used in psycholinguistic studies evaluating humans' sensitivity to argument roles, which allows us to use
carefully constructed minimal pairs of sentences which only differ with respect to argument roles, while controlling
for other factors like animacy. This serves as a rigorous test in examining models' ability to extract argument-role bindings based on sentence structure, as it requires models to go beyond simply learning relations between various arguments and verbs, i.e., between real-world events and participants that are likely to be involved in those events.
We compare model performance on two different types of argument role manipulations, in addition to a baseline condition which has shown to elicit immediate sensitivity in humans, as a way to more systematically compare human and model behavior.

Through three experiments, we find that i) language models show weak sensitivity to argument role information relative to role-independent argument meanings, similar to human initial prediction behavior, ii) models do not show the same consistency across different types of argument role manipulations as humans do, indicating a difference in the way argument roles are processed in models and humans, and iii) models' weak performance may not necessarily arise from inaccurate processing of argument roles. These results overall indicate that even if models are able to distinguish plausible and implausible verbs based on argument roles, to varying degrees of success, the lack of generalization across sentences that share the same structural relation suggests that the models do not use the same mechanism as humans to compute argument-verb relations. 

\section{Related Work}
To evaluate language models' representations of argument roles, reversing the order of the verb's arguments is a common design, paralleling the stimuli in human experiments.
Researchers then compare differences in the reversed and felicitous conditions, using various metrics from the models.
There are two major issues with existing work that we address.
First, existing work often relies on the animacy of the verbs' arguments.
Second, work using different metrics often offer conflicting conclusions.

\citet{papadimitriou-etal-2022-classifying-grammatical} claim language models are able to effectively make use of word order-related information when arguments are switched for verbs with transitive subjects and objects, reflecting these distinctions imposed by selectional constraints on the verb in their representations.
For instance, the models they evaluated would represent \textit{The chef chopped the onion} differently from \textit{The onion chopped the chef}.
For this evaluation, they automatically switch the order of arguments in naturalistic corpora.
Thus, it is unclear if these positive results are based on properties of the lexical items (i.e. frequency, animacy) that are learned more easily from distributional information, or more abstract representations of argument roles.\footnote{If such generalizations exist, they are largely tied to the presence of surface forms in the training data \cite{wilson-etal-2023-abstract}.
} 
A more reliable way to measure the linguistic capacity of language models is to effectively treat them as psycholinguistic subjects \citep[among others]{futrell-etal-2019-neural,ettinger-2020-bert} across a range of configurations (see reviews by \citet{linzen2021syntactic,pavlick2022semantic} and \citet{mahowald2024dissociating}).
Work in this vein presents models with minimal pairs of sentences and analyzes differences in language models' responses to each sentence.
Language models' sensitivity to a variety of phenomena been evaluated with this paradigm  \citep{linzen2016assessing,warstadt2020blimp,wilcox2023using}.
For argument roles specifically, \citet{kauf2023event} find they are able to distinguish plausible events from implausible ones, assigning higher probabilities to sentences like \textit{The teacher bought the laptop.} as opposed to \textit{The laptop bought the teacher.}, but only when one participant is animate and the other is inanimate.
Given the ability of language models to handle animacy even in atypical settings \citep{hannaetal2023language}, it is possible that the results of both \citet{kauf2023event} and \citet{papadimitriou-etal-2022-classifying-grammatical} may be tapping into this ability rather than a generalized representation of argument roles.

\citet{ettinger-2020-bert} presented a suite of psycholinguistically motivated diagnostics for BERT; one of these tests was on \textit{argument role reversals}, which was similar in spirit to some of \citet{kauf2023event}'s stimuli but only tested animate participants.
This study had different conclusions, finding that BERT was indeed sensitive to these role-related contrasts, generating role reversals in appropriate contexts, but not on par with humans.
Working with this dataset, \citet{li-etal-2021-bert} evaluate
the probabilities the models assign to the sentence at individual layers and finds that they are not sensitive to the role reversal sentences.
These studies all use different methods of evaluation. \citet{ettinger-2020-bert} queried sentence completions made by BERT, while \citet{kauf2023event} determined whether the language models assigned lower probabilities to the implausible sentence of the pair. 

We take a different approach to examine language models' sensitivity to argument roles by replicating
psycholinguistic experiments with multiple conditions designed to isolate humans' representations of argument roles.
These experiments track human processing in real time and specifically examine participants' responses to verbs, which reflect how the representation of the sentence is built up.
To tighten the link to whether models are making human-like judgments, we also examine the models' responses to the verbs rather than sentence-level metrics through behavioral and representational methods in Experiments 1 and 2.

Furthermore, one reason why Transformers are hypothesized to capture many empirical patterns in human sentence processing is that their attention mechanisms are able to efficiently keep track of long distance dependencies \citep{ryu-lewis-2021-accounting}.
Despite findings localizing handling certain syntactic dependencies to individual attention heads \citep{clark-etal-2019-bert,vig-belinkov-2019-analyzing,jian-reddy-2023-syntactic}, little work has been done on connecting these measures to psycholinguistic findings.
\citet{ryu-lewis-2021-accounting} specifically found an attention head that handled subject-verb agreement in GPT-2, which corresponded with human processing of these dependencies. 
This approach has not been tried for argument roles in a more generalized setting.\footnote{However, see improvements from \citet{timkey-linzen-2023-language} modeling this specific case and \citet{oh-schuler-2023-transformer} which shows the success of attention in modeling broad-coverage sentence processing.}
We do so in Experiment 3.
% \comment{NF}{I feel like you don't really close the loop here that you're going to fill this gap in the literature by connecting the attention analyses to psycholinguistic findings.  I wonder whether one way to do that would be to put the next paragraph before this one, and then start this paragraph with something like ``Furthermore, '' and end it with ``We do so here in Experiment 3'', or something like that.}

% \comment{NF}{This is a good paragraph to have here -- distinguish what makes your work different from previous work -- but the wording ended up a little jumbled.  I made some suggestions above, but if you have time to take another pass at it, I think it could still be improved.} \comment{SN}{I tried to focus on just evaluating the representation at the verb here.}

%by focusing on the representations of verbs appearing in role-appropriate and role-inappropriate contexts. We test language models' abilities to represent verbs in a role-specific way and the extent to which they distinguish the same verb when it appears in a contextually plausible and implausible context. We examine the robustness of this knowledge by comparing different instances of role-reversed contexts, where the role-appropriateness of the verb is manipulated in different ways. Finally, we compare how the models treat role-reversal sentences similarly or differently compared to other more lexically-driven changes in argument meanings.

\section{Psycholinguistic Data}
We use materials from previous psycholinguistic experiments which were carefully constructed to evaluate human comprehenders' sensitivity to argument roles in real-time sentence processing. These stimuli sets were designed to compare electrophysiological responses to verbs that appeared in different sentence contexts, and the different conditions have shown to elicit distinct N400 amplitudes, a neural response taken to reflect how strongly a target word was predicted based on the previous context \citep{kutas1980reading}.

We use the materials from \citet{chow2016bag} and \citet{kim2005independence}, and label the conditions as {\fontfamily{qcr}\selectfont swap-arguments}, {\fontfamily{qcr}\selectfont change-verb}, and {\fontfamily{qcr}\selectfont replace-argument} (Table \ref{stimuli_ex}).
Both studies were conducted in English on native speakers.

\begin{table*}[htbp]
\centering
\begin{tabular}{|p{3.75cm}|c| p{4.8cm}|p{4.8cm}|}
\hline
\textbf{Condition} & \textbf{Items} & \textbf{Plausible} & \textbf{Implausible} \\
\hline
{\fontfamily{qcr}\selectfont swap-arguments} & 120 & The restaurant owner forgot which \textit{customer} the \textit{waitress} \textbf{served} during dinner yesterday. & The restaurant owner forgot which \textit{waitress} the \textit{customer} \textbf{served} during dinner yesterday.\\ \hline
{\fontfamily{qcr}\selectfont change-verb} & 96 & The hearty meal was \textbf{devoured} with gusto. & The hearty meal was \textbf{devouring} by the kids. \\ \hline
{\fontfamily{qcr}\selectfont replace-argument} & 120 & The secretary confirmed which \textit{illustrator} the author had \textbf{hired} for the new book. & The secretary confirmed which \textit{readers} the author had \textbf{hired} for the new book. \\ \hline
\end{tabular}
\caption{Example sentences (1 pair = 1 item) in each condition. The {\fontfamily{qcr}\selectfont swap-arguments} and {\fontfamily{qcr}\selectfont change-verb} conditions involve argument role manipulations, while {\fontfamily{qcr}\selectfont replace-argument} serve as a control. Humans show greater sensitivity in the {\fontfamily{qcr}\selectfont replace-argument} than in the {\fontfamily{qcr}\selectfont swap-arguments} and {\fontfamily{qcr}\selectfont change-verb} conditions.}
\label{stimuli_ex} 
\end{table*}

Both the {\fontfamily{qcr}\selectfont swap-arguments} and {\fontfamily{qcr}\selectfont change-verb} conditions include manipulations of argument roles and verb plausibility. In the {\fontfamily{qcr}\selectfont swap-arguments} condition, the two arguments preceding the verb in the plausible sentence are swapped to create the implausible sentence. In the {\fontfamily{qcr}\selectfont change-verb} condition, the verb form is changed to create the plausible and implausible sentences. Although the two conditions involve different changes, both have the same consequence: verb plausibility changes because of the way the argument(s) are assigned different roles, while the argument(s) that appear in the context remain the the same (e.g., \textit{waitress-customer} or \textit{meal}).

In addition to the two role-related conditions, we also include a {\fontfamily{qcr}\selectfont replace-argument} condition \citep{chow2016bag}, which involves replacing one of the arguments with an entirely different noun. This results in changing the argument meaning rather than argument roles, and this has shown to yield immediate predictability effects in human verb predictions, as opposed to the previous two conditions which both fail to elicit rapid sensitivity.

% To summarize, 
The key human empirical pattern to which we compare language models' is: weaker sensitivity to argument roles ({\fontfamily{qcr}\selectfont swap-arguments} \& {\fontfamily{qcr}\selectfont change-verb}) compared to argument meanings ({\fontfamily{qcr}\selectfont replace-argument}).

\section{Models \& Experiments}
We use the following pre-trained language models for our analyses: GPT-2 (small, medium, and large) \citep{radford2019language}, BERT (base-uncased, large-uncased) \citep{devlin-etal-2019-bert}, and RoBERTa (base, large) \citep{liu2019roberta}. Details of the model properties are included in Appendix \ref{sec:appendix}.

These models were selected based on prior work comparing human language processing patterns with measures derived from language models.
Recent studies have shown that smaller versions of GPT-2 fit human reading times better than larger models \citep{oh-schuler-2023-surprisal, kuribayashi2023psychometric}. \citet{steuer-etal-2023-large} confirms these results, showing that larger Transformer language models perform better on syntactic and semantic generalization tasks than they do at predicting reading times relative to smaller models.
We selected different model sizes in order to examine how scaling up or down affects comparability with human performance.
Additionally, GPT-2 models are uni-directional while BERT models are bidirectional, but they have a similar number of parameters.
By manipulating the context available to a comprehender while controlling for model size, we can more effectively compare proxies of real-time incremental processing from the GPT-2 models compared to offline measures with the BERT-style models.

All models were accessed through the \texttt{transformers} \citep{wolf-etal-2020-transformers} or \texttt{minicons} library \citep{misra2022minicons}, built to work with the Huggingface API. Code and data are available at \href{https://github.com/umd-psycholing/RoleReversalLM}{\fontfamily{qcr}\selectfont https://github.com/umd-psycholing/RoleReversalLM}.

We carry out three experiments, evaluating language models' ability to differentiate plausible and implausible verbs given the sentence. We specifically focus on addressing the following questions: (i) Do the models show a human-like pattern across the different conditions? (ii) Are these contrasts reflected in the models' representations across the intermediate layers? (iii) Do patterns in the models' attention weights reflect argument role sensitivity? 

\section{Experiment 1: Surprisal Effects}
One of the most well-established measures linking language models to cognitive hypotheses is surprisal, or the negative log probability of a word given context.
Surprisal theory \citep{hale-2001-probabilistic,levy2008expectation} states that the difficulty associated with processing linguistic information can be operationalized with this measure.
Language model surprisal has shown to strongly correlate with both human reading times \cite{smith2013effect,shain2024large} as well as the N400 EEG response \cite{frank2013word,michaelov2024strong}.
Current Transformer models perform more effectively than other methods of language modeling \citep{merkx-frank-2021-human}, and this relationship with reading times has been established cross-linguistically \citep{wilcox-etal-2023-language}.
\subsection{Methods}
For each item, we compute the \textbf{surprisal effect} at the verb.
As human sensitivity to argument roles is often measured at the target verb, this allows us to make a direct comparison between humans and model-based measures of prediction.

Even if we might expect models to assign lower probability, and thus higher surprisal, to implausible continuations, it is important to determine the surprisal effect on individual items, following work on the targeted syntactic evaluation of language models \citep{marvin-linzen-2018-targeted,wilcox2023using}.
This allows us to quantify not just whether the model is successfully capturing distinctions between sentences, but to what extent it is able to do so.
We operationalize this effect in Equation \ref{surprisal_effect}, such that $context_i$ and $context_p$ are implausible and plausible versions of the same context, respectively, and $S_{LM}$ is the language model's surprisal in Equation \ref{surprisal_eqn}. 
\begin{equation} \label{surprisal_effect}
    S_{LM}(verb, context_i) - S_{LM}(verb, context_p)
\end{equation}
\begin{equation} \label{surprisal_eqn}
    S_{LM}(w, c) = - \log_2 P_{LM}(w | c)
\end{equation}

Verb surprisal estimates were obtained with Equation \ref{surprisal_effect}, and the surprisal effect for each item was obtained by subtracting the surprisal of the verb in the implausible context from the plausible context in all experimental conditions.
Therefore, a positive value indicates that the model correctly assigned lower surprisal to the target verb in the plausible context relative to the implausible context, i.e., role-sensitivity, while a value close to zero or negative indicates that the model incorrectly assigned similar or greater surprisal to the verb in the plausible context than the implausible context.

\subsection{Results}

\begin{figure*}[h!]
    \centering
\includegraphics[scale=.15]{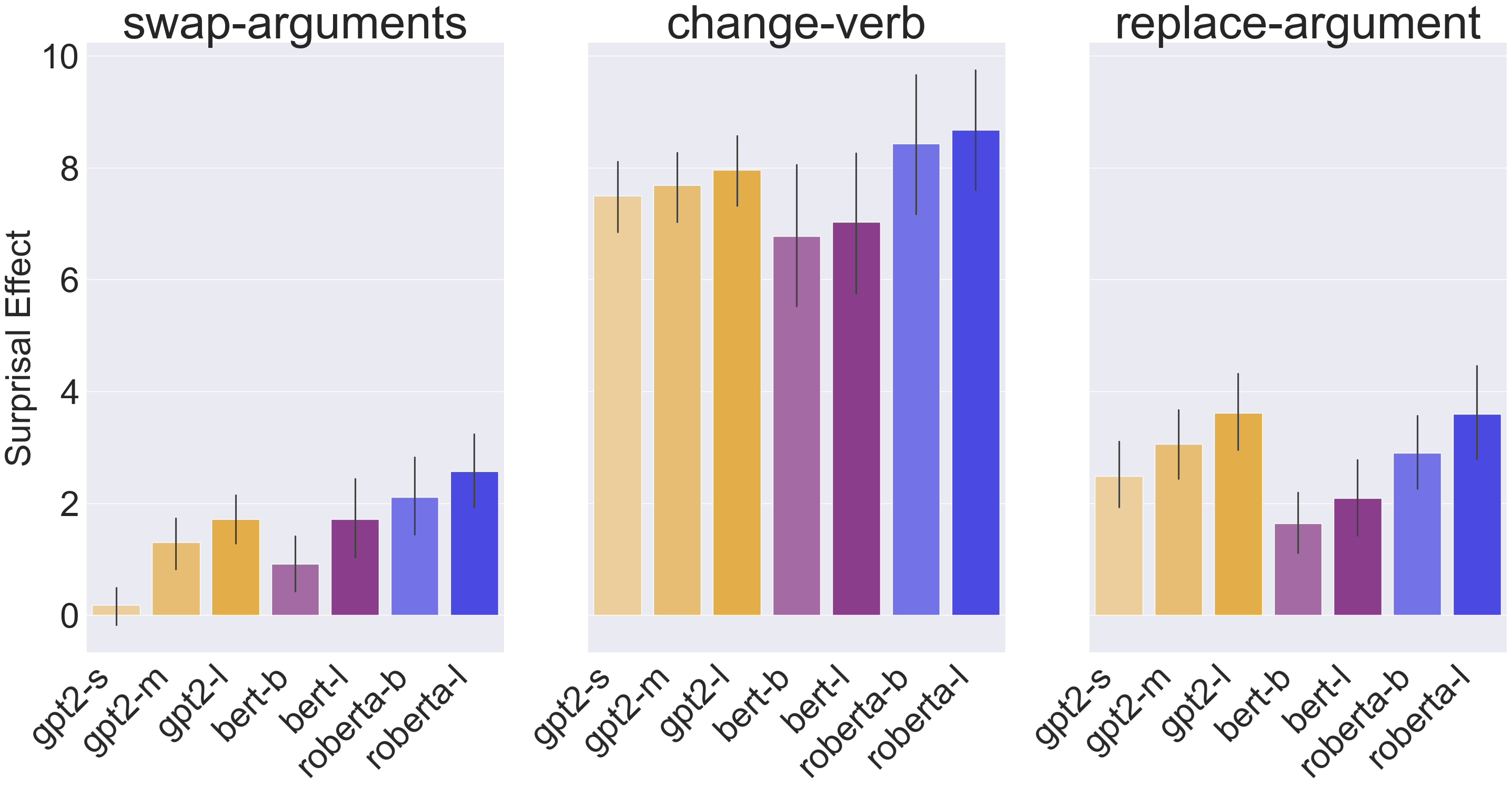}
    \caption{Surprisal effects plotted by condition and model. Higher values indicate greater role-sensitivity.} %.
    \label{fig:surprisal_effect}
\end{figure*}
We report the surprisal effect in all the models in Figure \ref{fig:surprisal_effect}.
In line with our expectations, the surprisal effect is larger for the {\fontfamily{qcr}\selectfont {\fontfamily{qcr}\selectfont replace-argument}} items than the {\fontfamily{qcr}\selectfont {\fontfamily{qcr}\selectfont swap-arguments}} items, showing that models are less sensitive to role reversals compared to {\fontfamily{qcr}\selectfont replace-argument}s. 
GPT2-small in particular did not exhibit any sensitivity to the role-reversed sentences, while showing considerably more sensitivity to the {\fontfamily{qcr}\selectfont replace-argument} sentences, consistent with \citet{chow2016bag}.
However, one key difference between the model and human responses is that all the models' effects for {\fontfamily{qcr}\selectfont change-verb} were far higher than both the {\fontfamily{qcr}\selectfont swap-arguments} and the baseline {\fontfamily{qcr}\selectfont replace-argument} case.
Instead of showing a smaller effect, like for {\fontfamily{qcr}\selectfont {\fontfamily{qcr}\selectfont swap-arguments}}, the surprisal effect for these sentences is far higher. 

The performance of GPT2-small for the {\fontfamily{qcr}\selectfont swap-arguments} condition mirrors the early stages of human processing more closely, as these role-reversed sentences do not elicit an N400 potential.
However, humans are also not sensitive to the manipulation in the {\fontfamily{qcr}\selectfont change-verb} stimuli since they use an abstract, generalized representation of argument roles, which is a major contrast with the models' surprisal.
Based on the comparably better performance on the {\fontfamily{qcr}\selectfont change-verb} and {\fontfamily{qcr}\selectfont replace-argument} conditions relative to {\fontfamily{qcr}\selectfont swap-arguments}, it is likely that the models are making use of specific lexical cues to make their inferences rather than the structural relations humans are using. 
This is because the two conditions the model does better on introduce lexical variation in the stimuli, which is not the case for {\fontfamily{qcr}\selectfont swap-arguments}.

\section{Experiment 2: Probing}
\subsection{Methods}
While the surprisal estimates in Experiment 1 are computed based on the final layer of the models, in Experiment 2, we investigate which layers encode argument role information in verb representations by conducting a probing analysis. To show role-sensitivity at the verb, the model must correctly analyze the position of the arguments, represent the arguments with a role-specific meaning, and use that information to determine the plausibility of the verb that appears following the arguments. As these computations involve both syntactic and semantic processing, it is possible that such knowledge is encoded in earlier layers of the models which are not detectable in surprisal estimates based on final layer representations \citep{tenney-etal-2019-bert,jawahar-etal-2019-bert}. We investigate this by implementing layer-wise \textit{probing classifiers} \citep{belinkov-2022-probing}, on GPT2-small, which showed the most human-like pattern in the surprisal analysis, as well as GPT2-medium, BERT-large, and RoBERTa-large, which have the same number of layers and show better performance with the {\fontfamily{qcr}\selectfont swap-arguments} condition than GPT2-small. 

For each condition, and for each layer, we train a logistic regression classifier on the models' representations of the target verbs, which predicts whether the verb is contextually appropriate or inappropriate.
We choose to use a linear classifier because evidence points to conceptually relevant information being linearly separable in embedding space \cite{nanda2023emergent}.
Target verbs in the plausible sentence were coded as 0 and the same target verbs in the implausible sentence were coded as 1.

Verb representations from each layer of each model were extracted using the \texttt{minicons} library. We report accuracies of each probe using 10-fold cross-validation with the \texttt{scikit-learn} implementation \citep{pedregosa2011scikit}. During  training, we used a controlled method of splitting the train and test data sets, where the plausible and implausible verb pairs were always included in the same data set. This was to prevent the model from simply matching a verb in one context to the same verb in the counterpart context.

A high \textbf{classification accuracy} indicates that the verb representations extracted from the model contains information about the plausibility of the verb given the sentence it appears in - the model is able to distinguish contextually appropriate and inappropriate verbs.

\subsection{Results}
\begin{figure*}[h!]
    \centering
    \includegraphics[width=\textwidth]{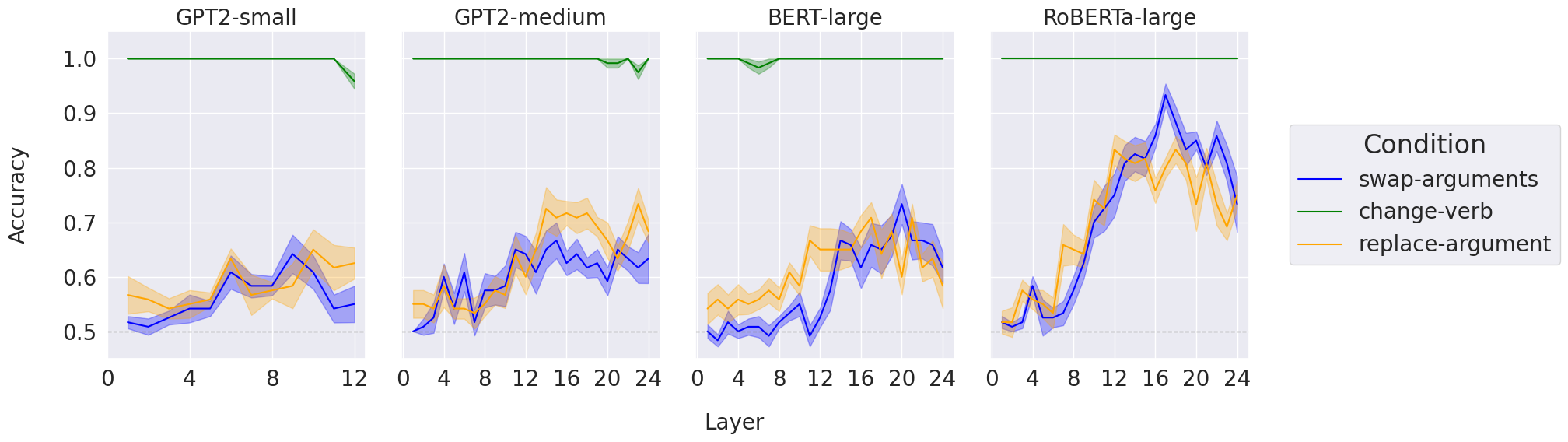}
    \caption{Classification accuracies for probes trained to distinguish plausible and implausible verbs under different conditions. Highlighted areas indicate standard errors of the mean across the 10 cross-validation folds. Dotted lines indicate at-chance accuracy.}
    \label{fig:probing_results}
\end{figure*}
The probes trained on the verb representations in the {\fontfamily{qcr}\selectfont change-verb} condition performed at ceiling for all models (Figure \ref{fig:probing_results}). This suggests that in all models, the systematic change in verb form (\textit{-ed} vs. \textit{-ing}) is robustly encoded in verb representations. This pattern corroborates the surprisal results, where the {\fontfamily{qcr}\selectfont change-verb} condition showed significantly large surprisal effects in all models, suggesting that the models can effectively distinguish verbs in the plausible and implausible contexts when the verb form differs between the two contexts.

Classification accuracy was generally lower for the conditions where the verb was kept the same and plausibility was determined by changing properties of the preceding context, i.e., {\fontfamily{qcr}\selectfont swap-arguments} \& {\fontfamily{qcr}\selectfont replace-argument}, rather than verb form. GPT2-small did not improve greatly from chance-level performance. The larger models reached higher classification accuracy, with GPT2-medium and BERT-large reaching 70\% accuracy, while RoBERTa showed the highest performance, reaching near 80-90\% accuracy. For these larger models, decoding accuracy gradually increased throughout the layers and the particular increase in the middle layers suggests that verb plausibility information is more effectively represented from the middle layers.

While the accuracies between the {\fontfamily{qcr}\selectfont swap-arguments} and {\fontfamily{qcr}\selectfont replace-argument} conditions were overall comparable, the {\fontfamily{qcr}\selectfont replace-argument} condition showed slightly higher accuracy than the {\fontfamily{qcr}\selectfont swap-arguments} condition in earlier layers of BERT and RoBERTa, while the same contrast appeared in later layers of GPT-2 (small and medium). This suggests that role-dependent verb plausibility information may be encoded at different stages of processing in uni- and bi-directional models.
Finally, there was a tendency for the accuracies to fluctuate more and even decrease at the final layers, particularly for the {\fontfamily{qcr}\selectfont swap-arguments} condition in RoBERTa, which drops from 90\% to 70\% accuracy. This suggests that role-dependent plausibility information may become partially lost in models' representations.

% \comment{NF}{What I'm really confused about here is how a two-alternative forced choice task could be significantly, and consistently, lower than 50\% accuracy for so many different models and layers.  What's going on there?  Are you sure there's not a bug in your code?}\comment{-------------RL}{There was indeed a bug (in some sense)! The models were really weirdly picking the exact opposite answer at first, but it turns out this was because the verb embeddings in plausible and implausible contexts are actually very similar, so when the models try to classify a plausible "served" (labeled 1) in the test data, for example, the most similar embedding in the training data would actually be the "served" in the implausible sentence (labeled 0), so they would incorrectly label it as implausible. I fixed this by ensuring that the same verb is always included in the same dataset when splitting into training and test.}

\section{Experiment 3: Attention}
\subsection{Methods}
One question based on the previous experiment findings is what gives rise to models' relatively weak performance on determining verb plausibility based on argument role information, particularly when the argument role is manipulated by swapping the position of the arguments ({\fontfamily{qcr}\selectfont swap-arguments} condition). One possibility is that for these items, the models often incorrectly parse the argument roles indicated by the structure. It is possible that the models get confused about which noun is in which position and takes on which argument role. This could also offer a reason for why models perform better with the {\fontfamily{qcr}\selectfont change-verb} items, where argument position is fixed and held constant between the plausible and implausible conditions. In Experiment 3, we examine how models treat the preceding arguments by conducting an attention analysis that focuses on whether the models correctly allocate attention to the target subject at the verb position.

We adapt a similar method to that used in previous work. \citet{ryu-lewis-2021-accounting} inspected the attention patterns of GPT-2 in order to probe whether the presence of a partially-matching distractor word interferes with the model's processing of a subject-verb dependency. The authors found an attention head that was specialized in finding the subject and examined whether the attention to the target subject differed between the intervening and non-intervening conditions. 

We compare the attention profiles of GPT2-small and RoBERTa-large, the models that performed the worst and best, respectively, in the previous experiments. For each model, we first define an attention head that allocates the greatest attention weight from the verb to the subject in the sentence. For example, given the sentence, \textit{The restaurant owner forgot which customer the waitress served during dinner yesterday}, we calculated the attention weight from the verb \textit{served} to the subject \textit{waitress} for each layer and head. We define the attention head that had the greatest attention weight to the subject as the subject attention head. The selected subject attention head was then used to calculate the attention from the verb to the subject and object, respectively. A high attention weight to the subject and a low attention weight to the object indicate that the model correctly distinguishes subjects from objects.

\begin{table*}[htbp]
\centering
\begin{tabular}{|c|c|c|c|c|c|}
    \hline
    \textbf{Model} & \textbf{Condition} & \multicolumn{2}{c|}{\textbf{Attention to Subject}} & \multicolumn{2}{c|}{\textbf{Attention to Object}} \\
    \cline{3-6}
    & & Plausible & Implausible & Plausible & Implausible \\
    \hline
    GPT2-small & {\fontfamily{qcr}\selectfont swap-arguments} & .53 (.15) & .53 (.17) & .18 (.10) & .19 (.06)\\
    \hline
    GPT2-small & {\fontfamily{qcr}\selectfont {\fontfamily{qcr}\selectfont replace-argument}}  & .51 (.12) & .50 (.13) & .19 (.09) & .21 (.08) \\
    \hline
    RoBERTa-large & {\fontfamily{qcr}\selectfont swap-arguments}  & .68 (.18) & .70 (.20) & .06 (.10) & .05 (.09)\\
    \hline
    RoBERTa-large & {\fontfamily{qcr}\selectfont {\fontfamily{qcr}\selectfont replace-argument}}  & .65 (.16) & .68 (.16) & .06 (.08) & .04 (.02) \\
    \hline
\end{tabular}
\caption{Results of the attention analysis. The values represent the subject attention head's average attention from the verb to the subject and its attention from the verb to the object under each condition. Standard deviations are in parentheses.}
\label{attention_results}
\end{table*}

\subsection{Results}
For GPT2-small, we identified layer 3 head 10 (head indices: 2, 9) as the subject attention head, and for RoBERTa-large, we identified layer 13 head 16 (head indicies: 12, 15) as the subject attention head. The attention weight to the subject averaged across all items was .52 for GPT2-small and .68 for RoBERTa, indicating that these attention heads allocated most of the attention from the verb to the subject across the experiment items.

The results are shown in Table \ref{attention_results}. We found similar attention patterns between the {\fontfamily{qcr}\selectfont swap-arguments} and {\fontfamily{qcr}\selectfont replace-argument} conditions. For both GPT-2 and RoBERTa, 
the subject attention head correctly allocates most of its attention to the subject rather than the object. However, RoBERTa gives less attention overall to the object than GPT-2 does, with the attention weight to the object remaining below 10\%. 

The results show that even GPT2-small, which did not show clear sensitivity to argument roles in the surprisal and probing analyses, correctly allocates attention to subjects with the subject head, though its attention is also distributed to the object more than the better performing RoBERTa-large. The attention analysis, therefore, suggests that it is unlikely that weak role-sensitivity at the verb arises from being confused about which argument is in which position or which argument is assigned which role. Rather, the weak performance could be due to how the models encode the preceding argument role information into the representations of the verb. Models may be able to correctly distinguish argument roles but less capable of using this information to represent role-compatible and role-inappropriate verbs in different ways.

\section{Discussion}
While previous studies have examined language models' knowledge of argument roles by testing their capacity to distinguish plausible and implausible sentences, we take a new approach by examining whether models' representations of verbs in sentences encode plausibility based on preceding argument role information. This method, in combination with the controlled sets of materials used in psycholinguistic studies that examine human comprehenders' role-sensitivity, offers a rigorous and systematic test of language models' sensitivity to argument roles and a way to directly compare human and model behavior. 
In the surprisal and probing analyses, we find that language models generally exhibit greater sensitivity to changes in argument meanings than to changes in argument roles, similar to humans' initial predictions.
However, unlike humans, they fail to show the same pattern across different types of argument role manipulations. Whether the argument role and verb compatibility is manipulated by swapping the argument positions or by changing the verb form, humans show the same processing pattern, whereas language models treat the two cases differently.

The relatively weak sensitivity to verb plausibility when the preceding arguments are swapped, which we observed in Experiments~1 and~2, is unlikely due to a misrepresentation of the context, as the models' attention patterns in Experiment~3 suggest that roles are accurately represented.
Rather, we suggest it arises from the difficulty in evaluating whether a verb is plausible given the particular argument-role bindings enforced by the preceding context. This involves a more complex analysis than simply computing context-independent argument and verb co-occurrences, which is potentially why humans' predictions fail to make use of such information rapidly during real-time prediction \cite{chow2016bag}.

A key divergence between the model and human behaviors was with regard to which conditions caused more difficulty than others. Human comprehenders show the same pattern in the {\fontfamily{qcr}\selectfont swap-arguments} and {\fontfamily{qcr}\selectfont change-verb} conditions (i.e., no immediate role-sensitivity), both of which involve determining a verb's fit with respect to given argument roles. In all the models we tested, we observed greater performance in the {\fontfamily{qcr}\selectfont change-verb} condition than the {\fontfamily{qcr}\selectfont swap-arguments} condition. This suggests that language models treat the two conditions differently, diverging from human processing behavior. The contrast between the role-related conditions further indicates that models do not compute argument-verb relations in those contexts using a shared underlying process, unlike human comprehenders who show similar role-sensitivity regardless of whether verb plausibility is manipulated through swapping the argument roles or changing the verb aspect. A possible explanation for this divergence between models and humans is that different morphological inflections of the same root could be represented as separate items in the language models' vocabulary (e.g., devouring - devoured), as opposed to how humans process variations in verb aspect. These results indicate that language models, like humans, may show differences in responses to plausible and implausible words or sentences, but the specific conditions under which these contrasts emerge can diverge (also see \citet{arehalli-etal-2022-syntactic,huang2024large}). This suggests that their performance may not rely on the same processing mechanisms as humans.

One notable observation was that GPT2-small showed stronger correspondence with the human N400 data patterns, while larger models showed the higher performance in all experiments, which outperformed humans' initial predictive processing capacities. 
GPT-2 and variants have shown to be more effective at predicting human behavior compared to larger autoregressive models \citep{oh2023does, kuribayashi2023psychometric}.
\citet{steuer-etal-2023-large} find a similar pattern, where smaller models predict human reading times better than larger ones that do better on syntactic and semantic judgments. Our results suggest that smaller models capture more immediate, online processing profiles of humans, and resemble human N400 patterns which reflect initial stages of predictive processing. Conversely, the measures derived from larger models more closely pattern with offline, final interpretations of humans. Nevertheless, no models capture the consistency between the two argument role manipulations which has been found with humans. These results offer insights into drawing connections with human empirical findings, especially for psycholinguists aiming to use language models, with regard to determining which models to use when simulating experiments. 
% \citet{michaelov2024strong} use GPT-3 to predict N400 amplitude, so if we replicate this experiment with GPT-3 surprisal and find the model is sensitive to role reversals, there will be an important empirical controversy to resolve, as GPT-2 surprisal may reflect the absence of an N400 effect more clearly than a larger model.
Additionally, the improved performance of larger models raises the question of whether scale is sufficient to learn these complex role-specific relationships; evaluating the argument role-reversal and {\fontfamily{qcr}\selectfont replace-argument} contrast for larger models like LLaMa \citep{touvron2023llama}, as well as tracing the ability based on the number of parameters of a language model, e.g., the Pythia family of models \cite{biderman2023pythia}, can facilitate these types of investigations.

Our work provides a critical perspective to language models' representations of argument roles from a psycholinguistic perspective.
Future directions could involve applying causal interpretability methods \citep{meng2022locating,arora2024causalgym} to these sets of sentences.
It may be the case that larger-scale models that assign correct plausibility ratings are implementing the similar computations for {\fontfamily{qcr}\selectfont replace-argument} and reversal items, which will take us further towards determining whether linguistic knowledge in language models is as robust as it seems.

\section*{Limitations}

\subsection*{Cross-Linguistic Coverage}
Our investigation was focused on English, but the role reversal effect has also been shown in languages like Mandarin \citep{chow2018wait} and German \citep{stonerole}.
Although it is linguistically robust across humans, \citet{xu2023linearity} found that language model surprisal exhibits different trends in each of these three languages.
Testing whether similar effects appear in other language models as well as monolingual or multilingual language models could be a way to establish whether the models' inferences are are based on language-specific factors or whether generalized representation of argument roles is an emergent phenomenon.

\subsection*{Interpretability}
Although it is unclear the extent to which attention-based measures provide explanatory value for model outputs on a variety of NLP tasks, a review from \citet{bibal-etal-2022-attention} suggests that the use of attention to explain syntactic parses is promising.
For our use case, attention heads that track dependencies are identified using correlational analyses, based on the weights between the verb and its arguments.
%Although this measure is easily understandable, it is likely that the attention heads do not just track subject-verb and object-verb dependencies. 
A key future direction is to build on work in interpretability \citep{lakretz2021mechanisms,meng2022locating} which identifies causal mechanisms in language models responsible for specific computations.
\citet{arora2024causalgym} apply some of these measures to pairs of grammatical and ungrammatical sentences handling various syntactic phenomena.
In future work, we hope to not just extend their methods, but derive measures of cognitive effort based on how the language models causally compute argument roles.

\section*{Ethical Considerations}
All data and language models we used were publicly available, and our experiments do not rely on any specialized computing hardware.

\section*{Acknowledgements}
We thank our reviewers, Colin Phillips, Navita Goyal, Rachel Rudinger, and other members of the Computational Cognitive Science group at UMD for providing feedback on this work. 
This work was supported by NSF grant DGE-2236417 and ONR MURI Award N00014-18-1-2670.
% Any opinions, findings, and conclusions or recommendations expressed in this material are those of the author(s) and do not necessarily reflect the views of the National Science Foundation.

\bibliography{anthology,custom}

\begin{thebibliography}{57}
\providecommand{\natexlab}[1]{#1}

\bibitem[{Arehalli et~al.(2022)Arehalli, Dillon, and Linzen}]{arehalli-etal-2022-syntactic}
Suhas Arehalli, Brian Dillon, and Tal Linzen. 2022.
\newblock \href {https://doi.org/10.18653/v1/2022.conll-1.20} {Syntactic surprisal from neural models predicts, but underestimates, human processing difficulty from syntactic ambiguities}.
\newblock In \emph{Proceedings of the 26th Conference on Computational Natural Language Learning (CoNLL)}, pages 301--313, Abu Dhabi, United Arab Emirates (Hybrid). Association for Computational Linguistics.

\bibitem[{Arora et~al.(2024)Arora, Jurafsky, and Potts}]{arora2024causalgym}
Aryaman Arora, Dan Jurafsky, and Christopher Potts. 2024.
\newblock Causalgym: Benchmarking causal interpretability methods on linguistic tasks.
\newblock \emph{arXiv preprint arXiv:2402.12560}.

\bibitem[{Belinkov(2022)}]{belinkov-2022-probing}
Yonatan Belinkov. 2022.
\newblock \href {https://doi.org/10.1162/coli_a_00422} {Probing classifiers: Promises, shortcomings, and advances}.
\newblock \emph{Computational Linguistics}, 48(1):207--219.

\bibitem[{Bibal et~al.(2022)Bibal, Cardon, Alfter, Wilkens, Wang, Fran{\c{c}}ois, and Watrin}]{bibal-etal-2022-attention}
Adrien Bibal, R{\'e}mi Cardon, David Alfter, Rodrigo Wilkens, Xiaoou Wang, Thomas Fran{\c{c}}ois, and Patrick Watrin. 2022.
\newblock \href {https://doi.org/10.18653/v1/2022.acl-long.269} {Is attention explanation? an introduction to the debate}.
\newblock In \emph{Proceedings of the 60th Annual Meeting of the Association for Computational Linguistics (Volume 1: Long Papers)}, pages 3889--3900, Dublin, Ireland. Association for Computational Linguistics.

\bibitem[{Biderman et~al.(2023)Biderman, Schoelkopf, Anthony, Bradley, O’Brien, Hallahan, Khan, Purohit, Prashanth, Raff et~al.}]{biderman2023pythia}
Stella Biderman, Hailey Schoelkopf, Quentin~Gregory Anthony, Herbie Bradley, Kyle O’Brien, Eric Hallahan, Mohammad~Aflah Khan, Shivanshu Purohit, USVSN~Sai Prashanth, Edward Raff, et~al. 2023.
\newblock Pythia: A suite for analyzing large language models across training and scaling.
\newblock In \emph{International Conference on Machine Learning}, pages 2397--2430. PMLR.

\bibitem[{Chow et~al.(2018)Chow, Lau, Wang, and Phillips}]{chow2018wait}
Wing-Yee Chow, Ellen Lau, Suiping Wang, and Colin Phillips. 2018.
\newblock Wait a second! delayed impact of argument roles on on-line verb prediction.
\newblock \emph{Language, Cognition and Neuroscience}, 33(7):803--828.

\bibitem[{Chow et~al.(2016)Chow, Smith, Lau, and Phillips}]{chow2016bag}
Wing-Yee Chow, Cybelle Smith, Ellen Lau, and Colin Phillips. 2016.
\newblock A “bag-of-arguments” mechanism for initial verb predictions.
\newblock \emph{Language, Cognition and Neuroscience}, 31(5):577--596.

\bibitem[{Clark et~al.(2019)Clark, Khandelwal, Levy, and Manning}]{clark-etal-2019-bert}
Kevin Clark, Urvashi Khandelwal, Omer Levy, and Christopher~D. Manning. 2019.
\newblock \href {https://doi.org/10.18653/v1/W19-4828} {What does {BERT} look at? an analysis of {BERT}{'}s attention}.
\newblock In \emph{Proceedings of the 2019 ACL Workshop BlackboxNLP: Analyzing and Interpreting Neural Networks for NLP}, pages 276--286, Florence, Italy. Association for Computational Linguistics.

\bibitem[{Devlin et~al.(2019)Devlin, Chang, Lee, and Toutanova}]{devlin-etal-2019-bert}
Jacob Devlin, Ming-Wei Chang, Kenton Lee, and Kristina Toutanova. 2019.
\newblock \href {https://doi.org/10.18653/v1/N19-1423} {{BERT}: Pre-training of deep bidirectional transformers for language understanding}.
\newblock In \emph{Proceedings of the 2019 Conference of the North {A}merican Chapter of the Association for Computational Linguistics: Human Language Technologies, Volume 1 (Long and Short Papers)}, pages 4171--4186, Minneapolis, Minnesota. Association for Computational Linguistics.

\bibitem[{Ettinger(2020)}]{ettinger-2020-bert}
Allyson Ettinger. 2020.
\newblock \href {https://doi.org/10.1162/tacl_a_00298} {What {BERT} is not: Lessons from a new suite of psycholinguistic diagnostics for language models}.
\newblock \emph{Transactions of the Association for Computational Linguistics}, 8:34--48.

\bibitem[{Frank et~al.(2013)Frank, Otten, Galli, and Vigliocco}]{frank2013word}
Stefan~L Frank, Leun~J Otten, Giulia Galli, and Gabriella Vigliocco. 2013.
\newblock Word surprisal predicts n400 amplitude during reading.

\bibitem[{Futrell et~al.(2019)Futrell, Wilcox, Morita, Qian, Ballesteros, and Levy}]{futrell-etal-2019-neural}
Richard Futrell, Ethan Wilcox, Takashi Morita, Peng Qian, Miguel Ballesteros, and Roger Levy. 2019.
\newblock \href {https://doi.org/10.18653/v1/N19-1004} {Neural language models as psycholinguistic subjects: Representations of syntactic state}.
\newblock In \emph{Proceedings of the 2019 Conference of the North {A}merican Chapter of the Association for Computational Linguistics: Human Language Technologies, Volume 1 (Long and Short Papers)}, pages 32--42, Minneapolis, Minnesota. Association for Computational Linguistics.

\bibitem[{Hale(2001)}]{hale-2001-probabilistic}
John Hale. 2001.
\newblock \href {https://aclanthology.org/N01-1021} {A probabilistic {E}arley parser as a psycholinguistic model}.
\newblock In \emph{Second Meeting of the North {A}merican Chapter of the Association for Computational Linguistics}.

\bibitem[{Hanna et~al.(2023)Hanna, Belinkov, and Pezzelle}]{hannaetal2023language}
Michael Hanna, Yonatan Belinkov, and Sandro Pezzelle. 2023.
\newblock \href {https://doi.org/10.18653/v1/2023.emnlp-main.744} {When language models fall in love: {A}nimacy processing in transformer language models}.
\newblock In \emph{Proceedings of the 2023 Conference on Empirical Methods in Natural Language Processing}, pages 12120--12135, Singapore. Association for Computational Linguistics.

\bibitem[{Huang et~al.(2024)Huang, Arehalli, Kugemoto, Muxica, Prasad, Dillon, and Linzen}]{huang2024large}
Kuan-Jung Huang, Suhas Arehalli, Mari Kugemoto, Christian Muxica, Grusha Prasad, Brian Dillon, and Tal Linzen. 2024.
\newblock Large-scale benchmark yields no evidence that language model surprisal explains syntactic disambiguation difficulty.
\newblock \emph{Journal of Memory and Language}, 137:104510.

\bibitem[{Jawahar et~al.(2019)Jawahar, Sagot, and Seddah}]{jawahar-etal-2019-bert}
Ganesh Jawahar, Beno{\^\i}t Sagot, and Djam{\'e} Seddah. 2019.
\newblock \href {https://doi.org/10.18653/v1/P19-1356} {What does {BERT} learn about the structure of language?}
\newblock In \emph{Proceedings of the 57th Annual Meeting of the Association for Computational Linguistics}, pages 3651--3657, Florence, Italy. Association for Computational Linguistics.

\bibitem[{Jian and Reddy(2023)}]{jian-reddy-2023-syntactic}
Jasper Jian and Siva Reddy. 2023.
\newblock \href {https://doi.org/10.18653/v1/2023.emnlp-main.144} {Syntactic substitutability as unsupervised dependency syntax}.
\newblock In \emph{Proceedings of the 2023 Conference on Empirical Methods in Natural Language Processing}, pages 2341--2360, Singapore. Association for Computational Linguistics.

\bibitem[{Kauf et~al.(2023)Kauf, Ivanova, Rambelli, Chersoni, She, Chowdhury, Fedorenko, and Lenci}]{kauf2023event}
Carina Kauf, Anna~A Ivanova, Giulia Rambelli, Emmanuele Chersoni, Jingyuan~Selena She, Zawad Chowdhury, Evelina Fedorenko, and Alessandro Lenci. 2023.
\newblock Event knowledge in large language models: the gap between the impossible and the unlikely.
\newblock \emph{Cognitive Science}, 47(11):e13386.

\bibitem[{Kim and Osterhout(2005)}]{kim2005independence}
Albert Kim and Lee Osterhout. 2005.
\newblock The independence of combinatory semantic processing: Evidence from event-related potentials.
\newblock \emph{Journal of memory and language}, 52(2):205--225.

\bibitem[{Kuribayashi et~al.(2023)Kuribayashi, Oseki, and Baldwin}]{kuribayashi2023psychometric}
Tatsuki Kuribayashi, Yohei Oseki, and Timothy Baldwin. 2023.
\newblock Psychometric predictive power of large language models.
\newblock \emph{arXiv preprint arXiv:2311.07484}.

\bibitem[{Kutas and Hillyard(1980)}]{kutas1980reading}
Marta Kutas and Steven~A Hillyard. 1980.
\newblock Reading senseless sentences: Brain potentials reflect semantic incongruity.
\newblock \emph{Science}, 207(4427):203--205.

\bibitem[{Lakretz et~al.(2021)Lakretz, Hupkes, Vergallito, Marelli, Baroni, and Dehaene}]{lakretz2021mechanisms}
Yair Lakretz, Dieuwke Hupkes, Alessandra Vergallito, Marco Marelli, Marco Baroni, and Stanislas Dehaene. 2021.
\newblock Mechanisms for handling nested dependencies in neural-network language models and humans.
\newblock \emph{Cognition}, 213:104699.

\bibitem[{Levy(2008)}]{levy2008expectation}
Roger Levy. 2008.
\newblock Expectation-based syntactic comprehension.
\newblock \emph{Cognition}, 106(3):1126--1177.

\bibitem[{Li et~al.(2021)Li, Zhu, Thomas, Xu, and Rudzicz}]{li-etal-2021-bert}
Bai Li, Zining Zhu, Guillaume Thomas, Yang Xu, and Frank Rudzicz. 2021.
\newblock \href {https://doi.org/10.18653/v1/2021.acl-long.325} {How is {BERT} surprised? layerwise detection of linguistic anomalies}.
\newblock In \emph{Proceedings of the 59th Annual Meeting of the Association for Computational Linguistics and the 11th International Joint Conference on Natural Language Processing (Volume 1: Long Papers)}, pages 4215--4228, Online. Association for Computational Linguistics.

\bibitem[{Linzen and Baroni(2021)}]{linzen2021syntactic}
Tal Linzen and Marco Baroni. 2021.
\newblock Syntactic structure from deep learning.
\newblock \emph{Annual Review of Linguistics}, 7:195--212.

\bibitem[{Linzen et~al.(2016)Linzen, Dupoux, and Goldberg}]{linzen2016assessing}
Tal Linzen, Emmanuel Dupoux, and Yoav Goldberg. 2016.
\newblock Assessing the ability of lstms to learn syntax-sensitive dependencies.
\newblock \emph{Transactions of the Association for Computational Linguistics}, 4:521--535.

\bibitem[{Liu et~al.(2019)Liu, Ott, Goyal, Du, Joshi, Chen, Levy, Lewis, Zettlemoyer, and Stoyanov}]{liu2019roberta}
Yinhan Liu, Myle Ott, Naman Goyal, Jingfei Du, Mandar Joshi, Danqi Chen, Omer Levy, Mike Lewis, Luke Zettlemoyer, and Veselin Stoyanov. 2019.
\newblock Roberta: A robustly optimized bert pretraining approach.
\newblock \emph{arXiv preprint arXiv:1907.11692}.

\bibitem[{Mahowald et~al.(2024)Mahowald, Ivanova, Blank, Kanwisher, Tenenbaum, and Fedorenko}]{mahowald2024dissociating}
Kyle Mahowald, Anna~A Ivanova, Idan~A Blank, Nancy Kanwisher, Joshua~B Tenenbaum, and Evelina Fedorenko. 2024.
\newblock Dissociating language and thought in large language models.
\newblock \emph{Trends in Cognitive Sciences}.

\bibitem[{Marvin and Linzen(2018)}]{marvin-linzen-2018-targeted}
Rebecca Marvin and Tal Linzen. 2018.
\newblock \href {https://doi.org/10.18653/v1/D18-1151} {Targeted syntactic evaluation of language models}.
\newblock In \emph{Proceedings of the 2018 Conference on Empirical Methods in Natural Language Processing}, pages 1192--1202, Brussels, Belgium. Association for Computational Linguistics.

\bibitem[{Meng et~al.(2022)Meng, Bau, Andonian, and Belinkov}]{meng2022locating}
Kevin Meng, David Bau, Alex Andonian, and Yonatan Belinkov. 2022.
\newblock Locating and editing factual associations in gpt.
\newblock \emph{Advances in Neural Information Processing Systems}, 35:17359--17372.

\bibitem[{Merkx and Frank(2021)}]{merkx-frank-2021-human}
Danny Merkx and Stefan~L. Frank. 2021.
\newblock \href {https://doi.org/10.18653/v1/2021.cmcl-1.2} {Human sentence processing: Recurrence or attention?}
\newblock In \emph{Proceedings of the Workshop on Cognitive Modeling and Computational Linguistics}, pages 12--22, Online. Association for Computational Linguistics.

\bibitem[{Michaelov et~al.(2024)Michaelov, Bardolph, Van~Petten, Bergen, and Coulson}]{michaelov2024strong}
James~A Michaelov, Megan~D Bardolph, Cyma~K Van~Petten, Benjamin~K Bergen, and Seana Coulson. 2024.
\newblock Strong prediction: Language model surprisal explains multiple n400 effects.
\newblock \emph{Neurobiology of language}, pages 1--29.

\bibitem[{Misra(2022)}]{misra2022minicons}
Kanishka Misra. 2022.
\newblock minicons: Enabling flexible behavioral and representational analyses of transformer language models.
\newblock \emph{arXiv preprint arXiv:2203.13112}.

\bibitem[{Nanda et~al.(2023)Nanda, Lee, and Wattenberg}]{nanda2023emergent}
Neel Nanda, Andrew Lee, and Martin Wattenberg. 2023.
\newblock Emergent linear representations in world models of self-supervised sequence models.
\newblock In \emph{Proceedings of the 6th BlackboxNLP Workshop: Analyzing and Interpreting Neural Networks for NLP}, pages 16--30.

\bibitem[{Oh and Schuler(2023{\natexlab{a}})}]{oh-schuler-2023-transformer}
Byung-Doh Oh and William Schuler. 2023{\natexlab{a}}.
\newblock \href {https://doi.org/10.18653/v1/2023.findings-emnlp.128} {Transformer-based language model surprisal predicts human reading times best with about two billion training tokens}.
\newblock In \emph{Findings of the Association for Computational Linguistics: EMNLP 2023}, pages 1915--1921, Singapore. Association for Computational Linguistics.

\bibitem[{Oh and Schuler(2023{\natexlab{b}})}]{oh-schuler-2023-surprisal}
Byung-Doh Oh and William Schuler. 2023{\natexlab{b}}.
\newblock \href {https://doi.org/10.1162/tacl_a_00548} {Why does surprisal from larger transformer-based language models provide a poorer fit to human reading times?}
\newblock \emph{Transactions of the Association for Computational Linguistics}, 11:336--350.

\bibitem[{Oh and Schuler(2023{\natexlab{c}})}]{oh2023does}
Byung-Doh Oh and William Schuler. 2023{\natexlab{c}}.
\newblock Why does surprisal from larger transformer-based language models provide a poorer fit to human reading times?
\newblock \emph{Transactions of the Association for Computational Linguistics}, 11:336--350.

\bibitem[{Papadimitriou et~al.(2022)Papadimitriou, Futrell, and Mahowald}]{papadimitriou-etal-2022-classifying-grammatical}
Isabel Papadimitriou, Richard Futrell, and Kyle Mahowald. 2022.
\newblock \href {https://doi.org/10.18653/v1/2022.acl-short.71} {When classifying grammatical role, {BERT} doesn{'}t care about word order... except when it matters}.
\newblock In \emph{Proceedings of the 60th Annual Meeting of the Association for Computational Linguistics (Volume 2: Short Papers)}, pages 636--643, Dublin, Ireland. Association for Computational Linguistics.

\bibitem[{Pavlick(2022)}]{pavlick2022semantic}
Ellie Pavlick. 2022.
\newblock Semantic structure in deep learning.
\newblock \emph{Annual Review of Linguistics}, 8:447--471.

\bibitem[{Pedregosa et~al.(2011)Pedregosa, Varoquaux, Gramfort, Michel, Thirion, Grisel, Blondel, Prettenhofer, Weiss, Dubourg et~al.}]{pedregosa2011scikit}
Fabian Pedregosa, Ga{\"e}l Varoquaux, Alexandre Gramfort, Vincent Michel, Bertrand Thirion, Olivier Grisel, Mathieu Blondel, Peter Prettenhofer, Ron Weiss, Vincent Dubourg, et~al. 2011.
\newblock Scikit-learn: Machine learning in python.
\newblock \emph{the Journal of machine Learning research}, 12:2825--2830.

\bibitem[{Radford et~al.(2019)Radford, Wu, Child, Luan, Amodei, Sutskever et~al.}]{radford2019language}
Alec Radford, Jeffrey Wu, Rewon Child, David Luan, Dario Amodei, Ilya Sutskever, et~al. 2019.
\newblock Language models are unsupervised multitask learners.
\newblock \emph{OpenAI blog}, 1(8):9.

\bibitem[{Ryu and Lewis(2021)}]{ryu-lewis-2021-accounting}
Soo~Hyun Ryu and Richard Lewis. 2021.
\newblock \href {https://doi.org/10.18653/v1/2021.cmcl-1.6} {Accounting for agreement phenomena in sentence comprehension with transformer language models: Effects of similarity-based interference on surprisal and attention}.
\newblock In \emph{Proceedings of the Workshop on Cognitive Modeling and Computational Linguistics}, pages 61--71, Online. Association for Computational Linguistics.

\bibitem[{Shain et~al.(2024)Shain, Meister, Pimentel, Cotterell, and Levy}]{shain2024large}
Cory Shain, Clara Meister, Tiago Pimentel, Ryan Cotterell, and Roger Levy. 2024.
\newblock Large-scale evidence for logarithmic effects of word predictability on reading time.
\newblock \emph{Proceedings of the National Academy of Sciences}, 121(10):e2307876121.

\bibitem[{Smith and Levy(2013)}]{smith2013effect}
Nathaniel~J Smith and Roger Levy. 2013.
\newblock The effect of word predictability on reading time is logarithmic.
\newblock \emph{Cognition}, 128(3):302--319.

\bibitem[{Steuer et~al.(2023)Steuer, Mosbach, and Klakow}]{steuer-etal-2023-large}
Julius Steuer, Marius Mosbach, and Dietrich Klakow. 2023.
\newblock \href {https://doi.org/10.18653/v1/2023.conll-babylm.12} {Large {GPT}-like models are bad babies: A closer look at the relationship between linguistic competence and psycholinguistic measures}.
\newblock In \emph{Proceedings of the BabyLM Challenge at the 27th Conference on Computational Natural Language Learning}, pages 142--157, Singapore. Association for Computational Linguistics.

\bibitem[{Stone and Rabovsky(2024)}]{stonerole}
Kate Stone and Milena Rabovsky. 2024.
\newblock The role of syntactic and semantic cues in preventing illusions of plausibility.

\bibitem[{Tenney et~al.(2019)Tenney, Das, and Pavlick}]{tenney-etal-2019-bert}
Ian Tenney, Dipanjan Das, and Ellie Pavlick. 2019.
\newblock \href {https://doi.org/10.18653/v1/P19-1452} {{BERT} rediscovers the classical {NLP} pipeline}.
\newblock In \emph{Proceedings of the 57th Annual Meeting of the Association for Computational Linguistics}, pages 4593--4601, Florence, Italy. Association for Computational Linguistics.

\bibitem[{Timkey and Linzen(2023)}]{timkey-linzen-2023-language}
William Timkey and Tal Linzen. 2023.
\newblock \href {https://doi.org/10.18653/v1/2023.findings-emnlp.582} {A language model with limited memory capacity captures interference in human sentence processing}.
\newblock In \emph{Findings of the Association for Computational Linguistics: EMNLP 2023}, pages 8705--8720, Singapore. Association for Computational Linguistics.

\bibitem[{Touvron et~al.(2023)Touvron, Martin, Stone, Albert, Almahairi, Babaei, Bashlykov, Batra, Bhargava, Bhosale et~al.}]{touvron2023llama}
Hugo Touvron, Louis Martin, Kevin Stone, Peter Albert, Amjad Almahairi, Yasmine Babaei, Nikolay Bashlykov, Soumya Batra, Prajjwal Bhargava, Shruti Bhosale, et~al. 2023.
\newblock Llama 2: Open foundation and fine-tuned chat models.
\newblock \emph{arXiv preprint arXiv:2307.09288}.

\bibitem[{Vig and Belinkov(2019)}]{vig-belinkov-2019-analyzing}
Jesse Vig and Yonatan Belinkov. 2019.
\newblock \href {https://doi.org/10.18653/v1/W19-4808} {Analyzing the structure of attention in a transformer language model}.
\newblock In \emph{Proceedings of the 2019 ACL Workshop BlackboxNLP: Analyzing and Interpreting Neural Networks for NLP}, pages 63--76, Florence, Italy. Association for Computational Linguistics.

\bibitem[{Warstadt et~al.(2020)Warstadt, Parrish, Liu, Mohananey, Peng, Wang, and Bowman}]{warstadt2020blimp}
Alex Warstadt, Alicia Parrish, Haokun Liu, Anhad Mohananey, Wei Peng, Sheng-Fu Wang, and Samuel~R Bowman. 2020.
\newblock Blimp: The benchmark of linguistic minimal pairs for english.
\newblock \emph{Transactions of the Association for Computational Linguistics}, 8:377--392.

\bibitem[{Wilcox et~al.(2023{\natexlab{a}})Wilcox, Meister, Cotterell, and Pimentel}]{wilcox-etal-2023-language}
Ethan Wilcox, Clara Meister, Ryan Cotterell, and Tiago Pimentel. 2023{\natexlab{a}}.
\newblock \href {https://doi.org/10.18653/v1/2023.emnlp-main.466} {Language model quality correlates with psychometric predictive power in multiple languages}.
\newblock In \emph{Proceedings of the 2023 Conference on Empirical Methods in Natural Language Processing}, pages 7503--7511, Singapore. Association for Computational Linguistics.

\bibitem[{Wilcox et~al.(2023{\natexlab{b}})Wilcox, Futrell, and Levy}]{wilcox2023using}
Ethan~Gotlieb Wilcox, Richard Futrell, and Roger Levy. 2023{\natexlab{b}}.
\newblock Using computational models to test syntactic learnability.
\newblock \emph{Linguistic Inquiry}, pages 1--44.

\bibitem[{Wilson et~al.(2023{\natexlab{a}})Wilson, Petty, and Frank}]{wilson2023abstract}
Michael Wilson, Jackson Petty, and Robert Frank. 2023{\natexlab{a}}.
\newblock How abstract is linguistic generalization in large language models? experiments with argument structure.
\newblock \emph{Transactions of the Association for Computational Linguistics}, 11:1377--1395.

\bibitem[{Wilson et~al.(2023{\natexlab{b}})Wilson, Petty, and Frank}]{wilson-etal-2023-abstract}
Michael Wilson, Jackson Petty, and Robert Frank. 2023{\natexlab{b}}.
\newblock \href {https://doi.org/10.1162/tacl_a_00608} {How abstract is linguistic generalization in large language models? experiments with argument structure}.
\newblock \emph{Transactions of the Association for Computational Linguistics}, 11:1377--1395.

\bibitem[{Wolf et~al.(2020)Wolf, Debut, Sanh, Chaumond, Delangue, Moi, Cistac, Rault, Louf, Funtowicz, Davison, Shleifer, von Platen, Ma, Jernite, Plu, Xu, Le~Scao, Gugger, Drame, Lhoest, and Rush}]{wolf-etal-2020-transformers}
Thomas Wolf, Lysandre Debut, Victor Sanh, Julien Chaumond, Clement Delangue, Anthony Moi, Pierric Cistac, Tim Rault, Remi Louf, Morgan Funtowicz, Joe Davison, Sam Shleifer, Patrick von Platen, Clara Ma, Yacine Jernite, Julien Plu, Canwen Xu, Teven Le~Scao, Sylvain Gugger, Mariama Drame, Quentin Lhoest, and Alexander Rush. 2020.
\newblock \href {https://doi.org/10.18653/v1/2020.emnlp-demos.6} {Transformers: State-of-the-art natural language processing}.
\newblock In \emph{Proceedings of the 2020 Conference on Empirical Methods in Natural Language Processing: System Demonstrations}, pages 38--45, Online. Association for Computational Linguistics.

\bibitem[{Xu et~al.(2023)Xu, Chon, Liu, and Futrell}]{xu2023linearity}
Weijie Xu, Jason Chon, Tianran Liu, and Richard Futrell. 2023.
\newblock The linearity of the effect of surprisal on reading times across languages.
\newblock In \emph{Findings of the Association for Computational Linguistics: EMNLP 2023}, pages 15711--15721.

\end{thebibliography}
%\bibliography{custom}

\appendix

\section{Computational Resources}
\label{sec:appendix}
Details of the model architectures we used are in Table \ref{tab:model_summary}.
All experiments were run on a single CPU and took no more than two hours to run.
We report metrics from a single run.

\begin{table}[ht]
\centering
\begin{tabular}{|c|c|c|c|c|c|c|c|}
    \hline
    \textbf{Model} & \textbf{Parameters} & \textbf{\#L} & \textbf{\#U} & \textbf{\#H} \\
    \hline
    GPT2 S & 124M & 12 & 768 & 12 \\
    \hline
    GPT2 M & 355M & 24 & 1024 & 16 \\
    \hline
    GPT2 L & 774M & 36 & 1280 & 20 \\
    \hline
    BERT B & 110M & 12 & 768 & 12 \\
    \hline
    BERT L & 340M & 24 & 1024 & 16 \\
    \hline
    RoBERTa B & 125M & 12 & 768 & 12 \\
    \hline
    RoBERTa L & 355M & 24 & 1024 & 16 \\
    \hline
\end{tabular}
\caption{Summary of Model Architectures. \#L, \#U, \#H each refers to the number of layers, hidden units, and attention heads.}
\label{tab:model_summary}
\end{table}

\section{Control Items}
\label{sec:appendix_control}

We further examined a set of items included in each study \cite{chow2016bag,kim2005independence}, where the plausibility of the verb was manipulated by simply replacing the target verb with another verb or associating the target verb with another argument. These materials have shown to elicit immediate neural responses in human comprehenders, indicating sensitivity to the likelihood of a target word appearing in a plausible context. High cloze conditions are listed first.

\begin{enumerate}
        \item[a.] Abby brushed her teeth after every \textbf{meal}/\textbf{game} and every snack. \citet{chow2016bag}.
        \item[b.] The [\textit{hungry boys}]/[\textit{dusty tabletops}] were \textbf{devouring} the plate of cookies when Jack arrived. \citet{kim2005independence}, adapted.
\end{enumerate}

\begin{figure}[H]
    \centering
    \includegraphics[scale=0.14]{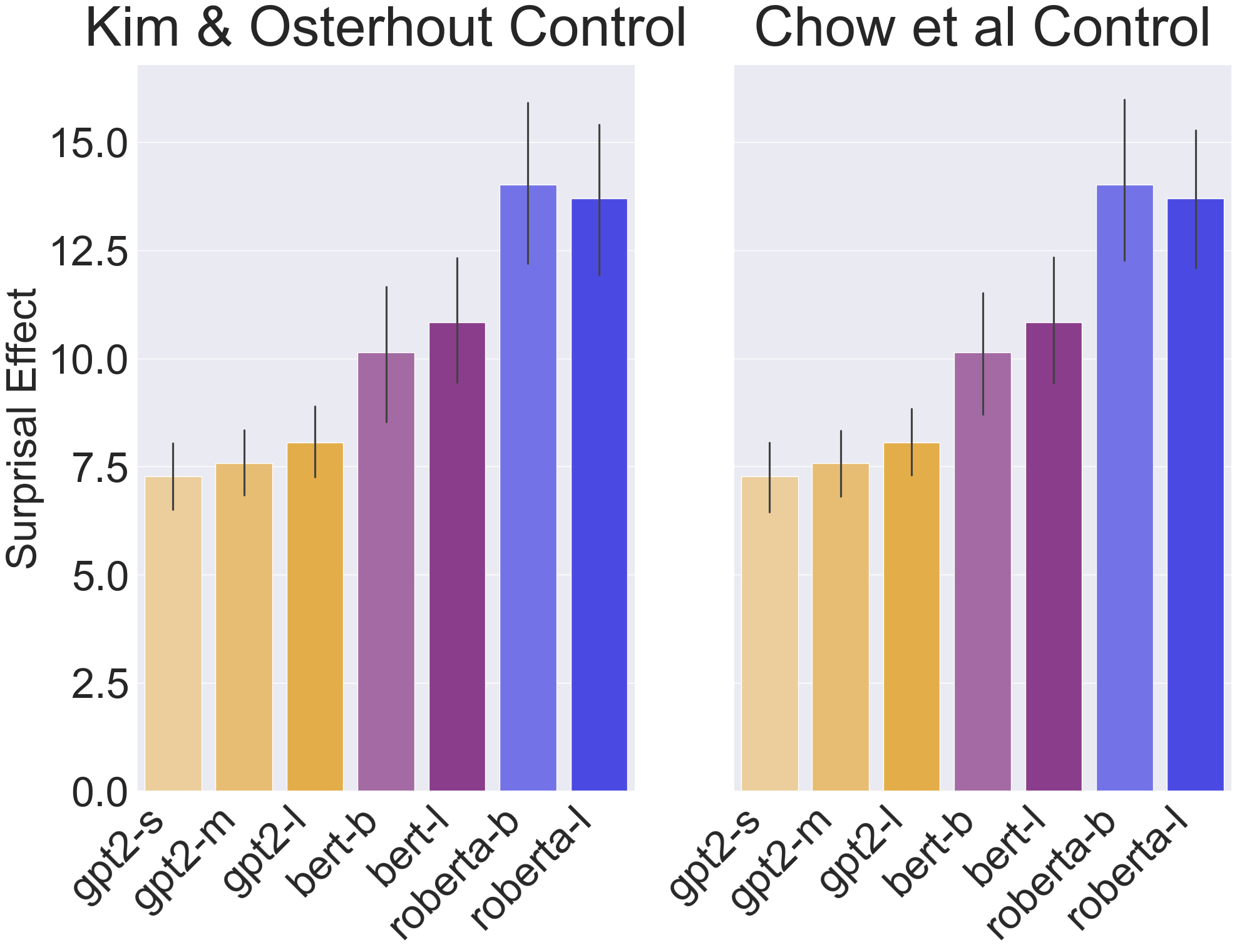}
    \caption{Surprisal effects for control items plotted by condition and model. Compare to {\fontfamily{qcr}\selectfont change-verb} for Kim \& Osterhout, {\fontfamily{qcr}\selectfont swap-arguments} and {\fontfamily{qcr}\selectfont replace-argument} for Chow et al.}
    \label{fig:fig3}
\end{figure}

We computed the surprisal effect for plausible and implausible variants of the same item for both studies, finding a much higher surprisal effect for both sets of control items (Figure \ref{fig:fig3}) relative to the experimental conditions (Figure \ref{fig:surprisal_effect}).

\appendix
\end{document}